# Intraoperative Organ Motion Models with an Ensemble of Conditional Generative Adversarial Networks

Yipeng Hu[1,2], Eli Gibson[1], Tom Vercauteren[1], Hashim U. Ahmed[3], Mark Emberton[3], Caroline M. Moore[3], J. Alison Noble[2], Dean C. Barratt[1]

[1] Centre for Medical Image Computing, University College London, London, UK
[2] Institute of Biomedical Engineering, University of Oxford, Oxford, UK
[3] Division of Surgery & Interventional Science, University College London, London, UK

**Abstract.** In this paper, we describe how a patient-specific, ultrasound-probe-induced prostate motion model can be directly generated from a single preoperative MR image. Our motion model allows for sampling from the conditional distribution of dense displacement fields, is encoded by a generative neural network conditioned on a medical image, and accepts random noise as additional input. The generative network is trained by a minimax optimisation with a second discriminative neural network, tasked to distinguish generated samples from training motion data. In this work, we propose that 1) jointly optimising a third conditioning neural network that pre-processes the input image, can effectively extract patient-specific features for conditioning; and 2) combining multiple generative models trained separately with heuristically pre-disjointed training data sets can adequately mitigate the problem of mode collapse. Trained with diagnostic T2-weighted MR images from 143 real patients and 73,216 3D dense displacement fields from finite element simulations of intraoperative prostate motion due to transrectal ultrasound probe pressure, the proposed models produced physically-plausible patient-specific motion of prostate glands. The ability to capture biomechanically simulated motion was evaluated using two errors representing generalisability and specificity of the model. The median values, calculated from a 10-fold cross-validation, were 2.8±0.3 mm and 1.7±0.1 mm, respectively. We conclude that the introduced approach demonstrates the feasibility of applying state-of-the-art machine learning algorithms to generate organ motion models from patient images, and shows significant promise for future research.

## 1    Introduction

Modelling patient-specific intraoperative organ motion provides biophysically informed constraints in many medical image computing tasks, such as anatomy tracking, segmentation, and multimodality image registration. In interventional applications, these tasks are crucial to aid intraoperative navigation and/or deliver a preoperative-image-based surgical plan. For example, previous studies have proposed patient-specific motion models for prostate that are built from segmented preoperative MR images and use finite element simulations as training motion data [1, 2]. The models were then applied to constrain MR-to-ultrasound registration algorithms to predict the MR-identified tumour location on transrectal ultrasound (TRUS) images, guiding targeted biopsy and focal therapies. It was reported that the motion models were robust to sparse and noisy data which are usually encountered in the intraoperative setting [1, 2].

Arguably, one of the most successful methods to summarise the organ motion distribution is based on principal component analysis (PCA), e.g. [1, 2, 3]. However, when applied to dense displacement fields (DDFs) that capture organ motion, this linear dimension reduction method 1) requires quality correspondence due to assumptions, such as orthogonality and importance of variance; and 2) is difficult to scale up to handle very large data sets. In computer-assisted intervention applications, subtle and often under-represented motion features, such as highly nonlinear local morphological change and pathological tissue deformation, are potentially relevant.

We propose an alternative method to model organ motion using generative neural networks that are both flexible enough to learn complex functions and have the potential to overcome both limitations. Such networks can be trained in an adversarial manner, as described in generative adversarial networks (GANs) [4]. If these networks are trained directly from multiple patient data sets, the resulting marginalised distribution represents motion across the entire population. However, in most surgical and interventional applications, *patient-specific* motion data is required. Approaches such as dynamic imaging and biophysical simulations are important sources of training data, but can be expensive or practically infeasible to obtain for every new patient [5, 6]. Therefore, we wish to estimate the conditional motion distribution given certain patient characteristics, such as a single preoperative (diagnostic or treatment planning) image, so that a patient-conditioned motion model can be inferred for a new patient directly from her/his medical image, i.e. without requiring additional patient-specific training data collection or model training for each new patient.

In this work, conditional GANs, e.g. [7, 8], are applied to model organ motion using medical images as conditioning data. We demonstrate the feasibility of building intraoperative prostate motion models from preoperative MR images, trained on the data set in the example prostate application described above [1, 2].

## 2    Patient-specific Organ Motion Model Generation

In this work, patient-specific organ motion is represented by DDFs, containing 3D displacement vectors uniformly sampled at 3D Cartesian coordinate grids of the preoperative image **y**. Given a patient image from a uniform prior distribution over an available preoperative image database $\mathbf{y} \sim P_{preop}(\mathbf{y})$; a training data set **x**, containing a set of DDFs (here, computed from finite element simulations as in [1, 2]), represent the conditional motion distribution $P_{motion}(\mathbf{x}|\mathbf{y})$. Given multiple patient data, both motion and preoperative image can be sampled from the joint training data distribution $(\mathbf{x}, \mathbf{y}) \sim P_{data}(\mathbf{x}, \mathbf{y}) = P_{motion}(\mathbf{x}|\mathbf{y})P_{pre}(\mathbf{y})$. Details of the data collection and normalisation used in this work are summarised in Section 4.

A conditional generative network $G(\mathbf{z}, \mathbf{y})$, the *generator*, is a structured probabilistic model with latent parameters $\mathbf{\theta}_G$, mapping independent unit Gaussian noise $\mathbf{z} \sim N(\mathbf{z})$ to the observed DDF space for each given **y**. The aim is to optimise the generator so that it is capable of generating motion samples similar to the training data $\mathbf{x} \sim P_{motion}(\mathbf{x}|\mathbf{y})$ by only sampling from $N(\mathbf{z})$ with a given preoperative image.

In a zero-sum minimax optimisation described in the GANs framework [4], the generator is optimised indirectly through the *discriminator*, a second $\mathbf{y}$-conditioned neural network $D(\mathbf{x}, \mathbf{y})$ with latent parameters $\boldsymbol{\theta}_D$, which is trained to distinguish the generated DDF samples $G(\mathbf{z}, \mathbf{y})$ from training data $\mathbf{x}$. The discriminator maximises a value function representing correct classification (i.e. $\mathbf{x}$ being classified as true and $G(\mathbf{z}, \mathbf{y})$ being classified as false), while only discriminator parameters $\boldsymbol{\theta}_D$ are trainable. Alternately, the generator parameters $\boldsymbol{\theta}_G$ are optimised by minimising the same value (or heuristically maximising the chance of $G(\mathbf{z}, \mathbf{y})$ being classified as true [4]). Once convergence is reached, the generator is expected to generate samples indistinguishable from training data. The cost functions for the conditioned generator and discriminator are given by:

$$J^{(G)} = -\frac{1}{2}\mathbb{E}_{\mathbf{z} \sim N, \mathbf{y} \sim P_{preop}} \log D(G(\mathbf{z}, \mathbf{y}), \mathbf{y}) + \frac{1}{2}\lambda \mathbb{E}_{(\mathbf{x},\mathbf{y}) \sim P_{data}, \mathbf{z} \sim N} \|G(\mathbf{z}, \mathbf{y}) - \mathbf{x}\|_2^2 \quad (1)$$

and

$$J^{(D)} = -\frac{1}{2}\mathbb{E}_{(\mathbf{x},\mathbf{y}) \sim P_{data}} \log D(\mathbf{x}, \mathbf{y}) - \frac{1}{2}\mathbb{E}_{\mathbf{z} \sim N, \mathbf{y} \sim P_{preop}} \log(1 - D(G(\mathbf{z}, \mathbf{y}), \mathbf{y})) \quad (2)$$

respectively, where $\mathbb{E}$ is the statistical expectation and $\lambda$ (set to 0.01 in this work) is a scalar hyper-parameter for the second mixing $L^2$ regularisation term [7] in Eq. (1).

**Conditioner network**
As the conditioning data are high dimensional image data, we propose to jointly optimise a third neural network, the *conditioner* $C(\mathbf{y})$ with latent parameters $\boldsymbol{\theta}_C$, to pre-process the given preoperative image. This is motivated by potential benefits from *parameter sharing* in which common conditioning features may be extracted efficiently by optimising only the conditioner parameters, as opposed to optimising both sets of parameters from the generator and the discriminator. However, optimising the conditioner parameters $\boldsymbol{\theta}_C$ directly by minimising either cost function, $J^{(G)}$ or $J^{(D)}$ (with $C(\mathbf{y})$ used in lieu of $\mathbf{y}$ for conditioning purposes), was neither efficient nor usually effective in practice, and was likely to produce overfitted conditioning features. This is probably caused by the received gradient updates to decrease one targeted cost function, in either scenario, which must backpropagate via both functions, inevitably increasing the other cost function. These are illustrated as the red and blue data flows in Fig. 1. Therefore, we define a separate conditioner cost function, independently optimising $\boldsymbol{\theta}_C$ to increase the conditioning ability of $C(\mathbf{y})$, by only maximising the chance of training motion data being classified as true (as shown in yellow in Fig. 1), as follows:

$$J^{(C)} = -\frac{1}{2}\mathbb{E}_{(\mathbf{x},\mathbf{y}) \sim P_{data}} \log D(\mathbf{x}, C(\mathbf{y})) \quad (3)$$

After replacing $\mathbf{y}$ with $C(\mathbf{y})$ in Eqs. (1) and (2), the parameters of the discriminator, the generator and the conditioner are alternately updated in each iteration of a minibatch gradient descent scheme, minimising the cost functions in Eqs. (1), (2) and (3), respec-tively. We note that it does not alter the original zero-sum minimax objective. The conditioner does not have a payoff strategy of its own, but rather provides an opportunity to extract explicit patient-specific features used for conditioning (see examples in Section 4); and may be benefited from *multitask learning*. In theory, the conditioner parameters could be absorbed into the generator and into the discriminator, whilst providing an effective regularisation strategy in training.

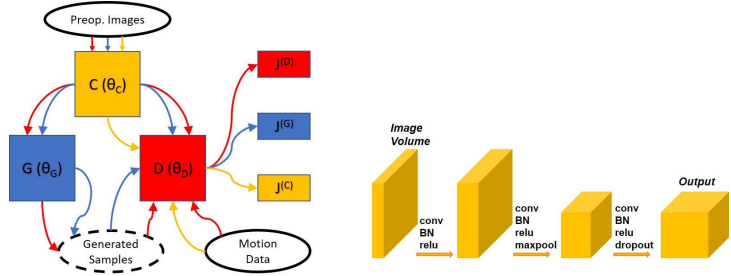

**Fig. 1.** Left: An illustration of three forward propagation paths in the proposed conditional GANs, where blue, red and yellow correspond to the generator, discriminator and conditioner, respectively. The contribution from the regularisation term in Eq. (1) is omitted here for clarity. Right: Illustration of the conditioner network (see text in Section 2 for details).

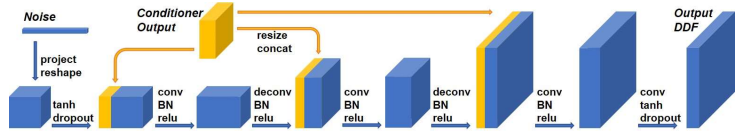

**Fig. 2.** Illustration of the generator network (see text in Section 2 for details).

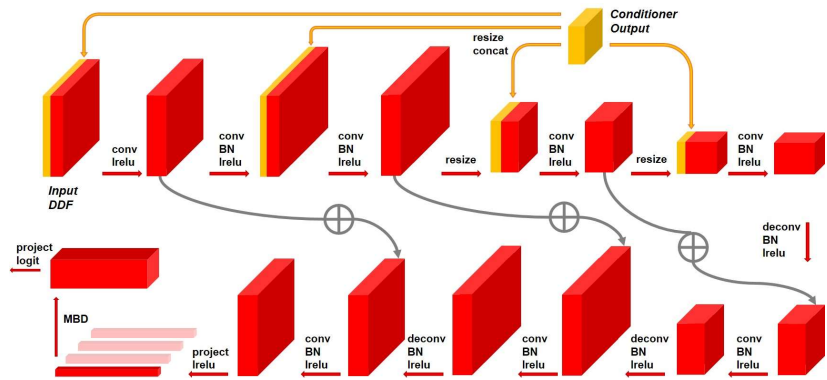

**Fig. 3.** Illustration of the discriminator network (see text in Section 2 for details).

**Network architecture**

The networks of the conditioner, the generator and the discriminator used in this study are illustrated in Figs. 1, 2 and 3, respectively. These were adapted from experience summarised in [9]. For simplicity, 3D medical images with channels and 4D DDFs are illustrated by 3D blocks.

The conditioner has three 3D convolutional (conv) layers with batch normalisation (BN) and rectified linear units (relu). In addition, maximum pooling (maxpool) and dropout (ratio=0.5) are added in the second and the third layers, respectively. Each layer doubles the number of channels of the previous layer, with 4 initial channels.

The generator takes input noise and propagates it forward through three pairs of 3D transposed convolutional (deconv) and conv layers mostly with BN and relu, up-sampling to the size of the DDFs. Exceptions are the first and the last layers, which use hyperbolic tangent function (tanh) as nonlinear activations and dropouts. Each sampling level halves the number of channels of the previous level, with 16 initial channels, and concatenates (concat) a properly resized conditioner output.

The discriminator has 32 initial channels and utilises two down-sampling layers by resizing, followed by up-sampling deconv-conv layers with leaky relu (lrelu). Three "residual network" style summation (circled plus sign) operations shortcut the network. A minibatch discrimination (MBD) layer [10], having 300 projected features with five 100 dimensional kernels, is also added to measure a relative distance within a minibatch of size 16, before the output logit layer with one-sided label smoothing [10]. The conditioner output is only concatenated with the down-sampling part of the network.

## 3   Combining Generative Adversarial Networks

A well-recognised problem in applying GANs is mode collapse - a phenomenon in which some part of the training data manifold is not represented in the trained generative model [9]. Although the minibatch discrimination layer [10] in the discriminator noticeably helped the generator to produce more diverse samples, mode collapse was still observed in our initial experiments. The motion samples produced by the converged conditional GANs were specific to individual patients and physically plausible, but it lacked coverage of the entire data manifold as confirmed by large errors representing poor generalisability (details discussed in Section 4).

For our application, we implemented a pragmatic approach to simplify the objective for the generator, in which the training data were partitioned according to the sign of the average displacement in three dimensions, resulting in eight different data groups. An ensemble of eight conditional GANs was then trained with these pre-clustered motion data independently. Each of these generative models can be considered to represent the distribution of a subpopulation, labelled by the data group $m = 1, ..., M$ (here $M = 8$). Therefore, each can generate samples conditioned on the group label. The original population distribution then can be recovered by a trivial model-averaging of these subpopulation distributions, marginalising over group label priors:

$$P_{motion}(x|y) = \sum_{m=1}^{M} P_{motion}(x|y, m) P(m) \qquad (4)$$

where the prior $P(m)$ can be estimated simply as fractions of the disjointed training data. When sampling, each motion sample has a $P(m)$ chance to be drawn from $m^{th}$ generator trained using $m^{th}$ group of data.

## 4    Experiments and Results

T2-weighted MR images were acquired from 143 patients who underwent TRUS-guided transperineal targeted biopsy or focal therapy for prostate cancer. The preoperative MR images were normalised to 2 × 2 × 2 mm/voxel with unit-variance and zero-mean intensity values. For each patient, 512 finite element (FE) simulations were performed using NiftySim (niftk.org) on a NVIDIA® GeForce™ 8600GT GPU, resulting in 72,216 simulated motion data. Each DDF was sampled from each simulation that predicts one plausible prostate motion due to change of ultrasound probe movement, acoustic balloon dilation and mechanical properties of soft tissues, subject to nearby pelvic bony constraints. Further details and validation of the simulations were described in previous studies [1, 2]. These DDFs were normalised to 5 × 5 × 5 mm/grid with a displacement range of [−1,1], and spatially aligned to common physical coordinates of the corresponding MR images.

The proposed models were implemented with TensorFlow™ and trained on a 12GB NVIDIA® Tesla™ K40 GPU, using the Adam optimiser with 100 unit Gaussian noise as prior input while an $L^2$ weight decay was set to 0.01. Random affine transformation was applied on the DDFs and the MR images in each iteration for data augmentation. Pre-clustering (Section 3) resulted in 3100-14536 training data over the eight groups.

A 10-fold cross-validation was performed: MR images and motion data from 14-15 patients were left-out as a validation set; For each patient in the validation set, 10,000 samples were generated from the model that was trained with the remaining training set; The *network-generated samples* were then compared with the left-out 512 *FE test data* (simulations from the same patient). We adapted a simple yet clinically relevant evaluation method that compares coverages of distribution support and can be related to target registration error and tumour detection rate [2], measuring the model generalisability and specificity in terms of root-mean-square errors (RMSEs) in displacement difference between network-generated samples and FE test data over entire DDF. The generalisability measures the network ability to generate all motions observed in the FE test data, defined as an average RMSE between all FE test data and their nearest network-generated samples; The specificity measures how specifically the network generates only plausible samples (i.e. similar to FE test data), defined as an average RMSE between all network-generated samples and the closest FE test data.

Without conditioning on patient images, for instance, a sampled DDF may not be spatially aligned with the patient anatomy, so that the prostate gland may incorrectly be in regions containing locally sharp displacement (possibly due to rigid probe movement), or implausible force may be exerted from the anterior, both resulting in unrealistic prediction of organ motion. The first row in Fig. 4 illustrates typical examples from unconditioned GANs, while samples generated by the proposed networks are also demonstrated in Fig. 4. These motion samples are visually plausible, diverse, and retained highly nonlinear local deformation often found near the posterior of the gland.

The overall median generalisability and specificity are 2.83±0.26 mm and 1.66±0.06 mm, respectively, which compare well with the results of a previous study, in which PCA was used [6]. The generalisabilities are significantly better (smaller RMSEs) compared to using single GANs (an implementation without pre-clustering training data described in Section 3), and significantly better specificities were found compared to using unconditioned GANs (a model trained without feeding preoperative images or the conditioner), for all 143 patients (for all $p<0.001$, paired t-tests at $\alpha=0.05$). Example individual results are plotted on the left in Fig. 5. On the right of Fig. 5, examples of the trained conditioner output are shown. Most interestingly, anatomical structures, such as prostate (indicated with green arrows) and the rectum (indicated with orange arrows), which were assigned different material properties or boundary conditions in the biomechanical models, are outlined effectively. This suggests that the conditioner may be extracting patient-specific anatomical information from the images.

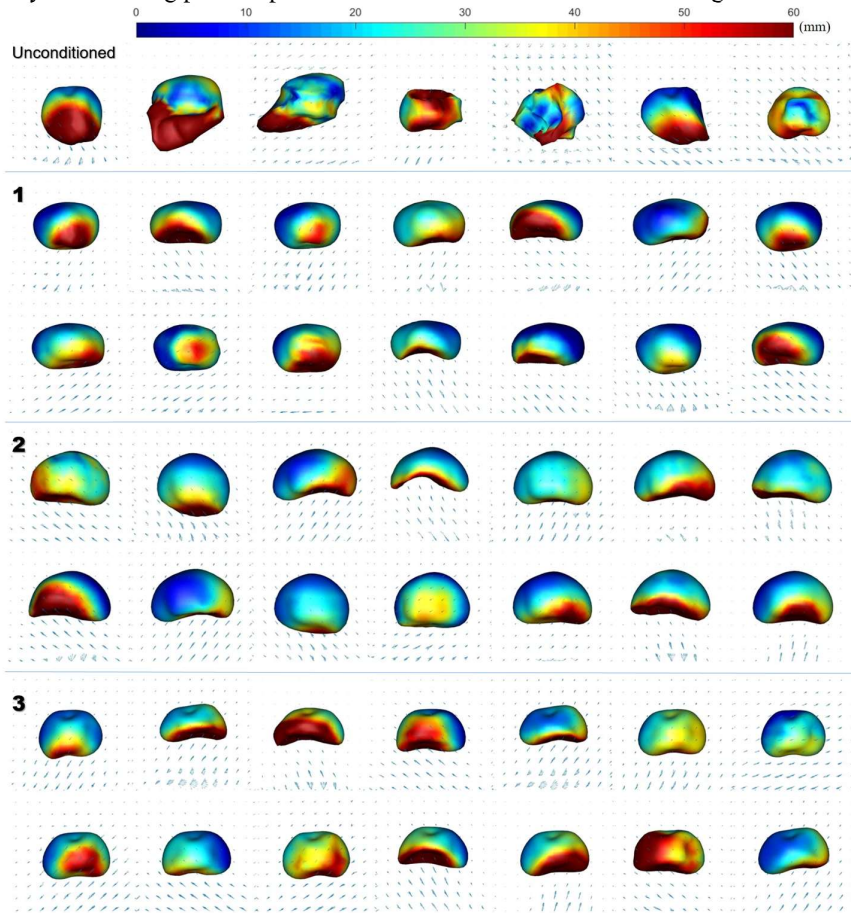

**Fig. 4.** The first row contains example DDFs sampled from unconditioned GANs; the remaining Subplots 1, 2 and 3 are example DDFs sampled from the proposed ensemble of conditional GANs, conditioned on three patient images. The prostate gland motion (colour-coded with displacement magnitude) was interpolated from the DDFs, indicated by blue arrows (cropped and coarsely-resampled for illustration purpose).

## 5 Discussion

In this paper, we report promising results of applying GANs, conditioned on preoperative images, to model patient-specific organ motion in a prostate cancer intervention application, and describe a strategy for overcoming the practical issue of mode collapse. While evaluating distribution of motion remains challenging, the proposed method offers several advantages over previous motion models: it can readily be trained on a large data set and can generate samples quickly; it can learn highly complex motion directly from medical image without segmentation, correspondence or other patient-specific information such as tissue properties. These may be important for many challenging applications, such as modelling pathological tissue motion, which we intend to investigate. Future research also includes improving model architectures and further investigating mode collapse to improve training efficiency and modelling ability.

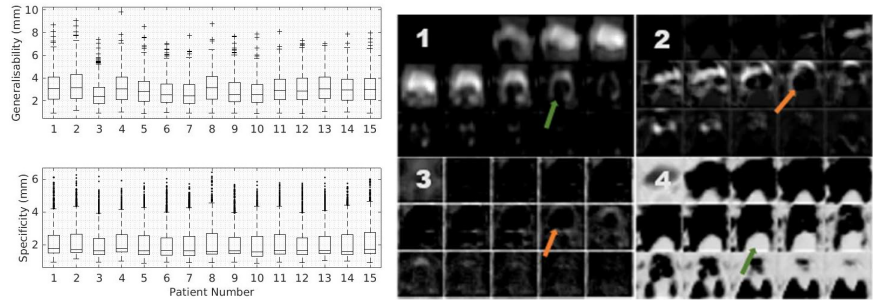

**Fig. 5.** Left: boxplots of the RMSEs from cross-validation representing generalisability (upper) and specificity (lower), defined in Section 4. Right: example montages of the conditioner output channels (1-4, cropped for illustration). See text in Section 4 for details.

**Acknowledgement:** This work is supported by the EPSRC and CRUK through the Comprehensive Cancer Imaging Centre (CCIC); Y. Hu also receives support from a CMIC Platform Fellowship (EP/M020533/1); T. Vercauteren and D. C. Barratt are also with the Wellcome / EPSRC Centre for Interventional and Surgical Sciences (WEISS). The data used in this study were collected during the SmartTarget® trials funded by the Wellcome Trust.

**Reference**
1. Wang, Y. et al., 2016. Towards personalized statistical deformable model and hybrid Point matching for robust MR-TRUS registration. *IEEE-TMI*, 35(2), 589-604.
2. Hu, Y. et al., 2012. MR to ultrasound registration for image-guided prostate interventions. *Medical image analysis*, 16(3), 687-703.
3. Rueckert, D. et al., 2003. Automatic construction of 3-D statistical deformation models of the brain using nonrigid registration. *IEEE-TMI*, 22(8), 1014-1025.
4. Goodfellow, I., et al., 2014. Generative adversarial nets. *NIPS 2014,* 2672-2680.
5. Zhu, Y. et al., 2010. Segmentation of the left ventricle from cardiac MR images using a subject-specific dynamical model. *IEEE-TMI*, 29(3), 669-687.

6. Hu, Y. et al., 2015. Population-based prediction of subject-specific prostate deformation for MR-to-ultrasound image registration. *Medical image analysis*, *26*(1), 332-344.
7. Pathak, D., et al., 2016. Context encoders: Feature learning by inpainting. In *CVPR 2016*, 2536-2544.
8. Gauthier, J., 2014. Conditional generative adversarial nets for convolutional face generation. *Stanford University Class Project Report, CS231N: CNNVR 2014*, 5.
9. Goodfellow, I., 2016. NIPS 2016 Tutorial: Generative Adversarial Networks. *arXiv preprint arXiv:1701.00160*.
10. Salimans, T. et al., 2016. Improved techniques for training gans. In *NIPS 2016,* 2226-2234.